\def\year{2023}\relax
\documentclass[letterpaper]{article} 
\usepackage{aaai23}  
\usepackage{times}  
\usepackage{helvet}  
\usepackage{courier}  
\usepackage[hyphens]{url}  
\usepackage{graphicx} 
\usepackage{natbib}  
\usepackage{caption} 
\DeclareCaptionStyle{ruled}{labelfont=normalfont,labelsep=colon,strut=off} 
\usepackage{algorithm}
\usepackage{algorithmic}

\usepackage{multirow}
\usepackage{subfigure} 
\usepackage{lipsum}
\usepackage{xcolor}
\usepackage{booktabs}
\usepackage{graphics}
\usepackage{amssymb}
\usepackage{amsmath}
\usepackage{pifont}
\usepackage{makecell}
\usepackage{wrapfig} 
\usepackage{booktabs}
\usepackage{fdsymbol}

\usepackage{tikz}
\usepackage{pgfplots}

\usepackage{newfloat}
\usepackage{listings}
\floatstyle{ruled}
\newfloat{listing}{tb}{lst}{}
\floatname{listing}{Listing}
\definecolor{bronze}{rgb}{0.0, 0.42, 0.24}

\newcommand{\taskname}[1]{G-PlanET} 
\newcommand{\metricname}[1]{KAS}
\newcommand{\eg}[1]{\textit{e.g.,}} 
\newcommand{\ie}[1]{\textit{i.e.,}} 

\title{
\vspace*{-0.5in}
{{\small \hfill  AAAI 2023}\\
\vspace*{.25in}}

On Grounded Planning for Embodied Tasks with Language Models}

\author{Bill Yuchen Lin\textsuperscript{{1}}\thanks{~The first two authors contributed equally.}, Chengsong Huang\textsuperscript{{2}}\footnotemark[1], Qian Liu\textsuperscript{{3}}, 
{{Wenda Gu}\textsuperscript{{1}}, {Sam Sommerer}\textsuperscript{{1}}, {Xiang Ren}\textsuperscript{{1}}}
}
\affiliations{
    \textsuperscript{\rm 1} University of Southern California \\
    \textsuperscript{\rm 2} Fudan University \quad  \textsuperscript{\rm 3}Sea AI Lab \\
   \{yuchen.lin, wendagu, sommerer, xiangren\}@usc.edu, huangcs19@fudan.edu.cn, liuqian@sea.com\\
}

\usepackage{bibentry}

\begin{document}

\maketitle

\begin{abstract}
Language models (LMs) have demonstrated their capability in possessing commonsense knowledge of the physical world, a crucial aspect of performing tasks in everyday life. However, it remains unclear whether they have the capacity to generate grounded, executable plans for embodied tasks. This is a challenging task as LMs lack the ability to perceive the environment through vision and feedback from the physical environment.
In this paper, 
we address this important research question and present the first investigation into the topic. Our novel problem formulation, named ~\taskname{}, inputs a high-level goal and a data table about objects in a specific environment, and then outputs a step-by-step actionable plan for a robotic agent to follow.
To facilitate the study, we establish an evaluation protocol and design a dedicated metric, \metricname{}, to assess the quality of the plans.
Our experiments demonstrate that the use of tables for encoding the environment and an iterative decoding strategy can significantly enhance the LMs' ability in grounded planning. Our analysis also reveals interesting and non-trivial findings. \footnote{Project website: \color{blue}{\underline{\textit{https://yuchenlin.xyz/g-planet/}}}}


\end{abstract}

\section{Introduction}
Pre-trained language models (LMs) demonstrate exceptional proficiency in a wide range of natural language processing (NLP) tasks such as question answering, machine translation, and summarization. They indeed capture some commonsense knowledge about our physical world such as ``birds can fly''. However, the question of whether LMs can exhibit reasoning abilities within a grounded, realistic setting remains an open issue. This is because LMs lack the sensory experiences and physical interactions with the environment that enable human beings to grasp the nuances of real-life situations and plan for completing tasks.
\begin{figure}[th]
\centering
\includegraphics[width=0.95\columnwidth]{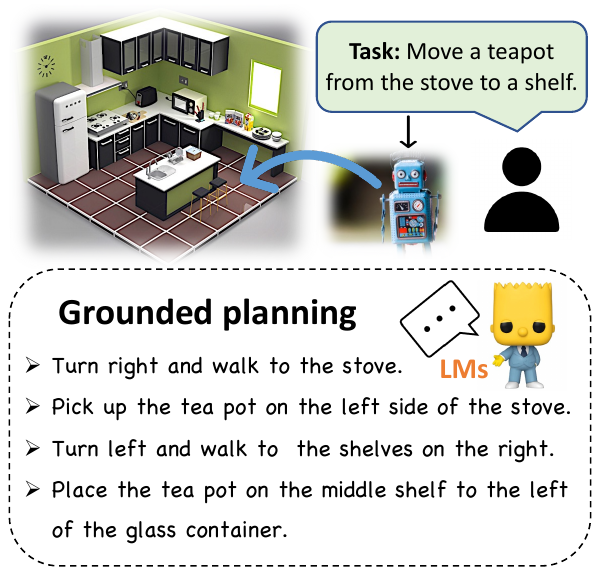}
\caption{The task of grounded planning for embodied tasks (\taskname{}). The input to the LMs is a goal with a specific environment, and the output is a step-by-step plan that can guide a robot to complete the task.}
\label{fig:intro}
\end{figure}

Embodied robotics learning is a growing field that seeks to create artificial intelligence agents capable of navigating and performing tasks within real-world environments, typically simulated through physical engines such as AI2THOR~\cite{ai2thor}. The ALFRED benchmark~\cite{ALFRED20} represents one of the pioneering datasets that bridges the gap between NLP and robotics, providing a platform for investigating language-directed agents. The objective of these studies is to design and test agents that can translate language instructions into sequences of low-level actions that enable the agent to manipulate objects within an environment and achieve a desired outcome (\eg{} cleaning an object and placing it elsewhere).


However, the primary emphasis of the ALFRED benchmark and related datasets is on the comprehension of pre-established plans, rather than the ability to reason and independently plan within a realistic environment. Prior research focuses on the capacity of agents to comprehend and execute step-by-step plans, but not on their capacity for decomposing tasks and generating such plans, which represents a more advanced skill. 
Additionally, the role of LMs has received limited examination in the context of these benchmarks, where they are mainly used as encoders for embedding token sequences, rather than for planning or reasoning.


Prior studies have explored the planning capability of LMs, with \citet{Huang2022LanguageMA} demonstrating that GPT-3 and similar models are capable of generating general plans for executing everyday tasks. However, these plans lack grounding in a realistic environment, as LMs are not environment-specific. As a result, these plans are not necessarily executable by agents. For instance, in the context of an ALFRED task to ``move a teapot from the stove to a shelf,'' embodied agents require knowledge of the location of the teapot and the path to reach it.  Humans, on the other hand, can readily observe the location of the teapot on the stove and their current position in the kitchen, allowing them to formulate a grounded plan that starts with ``turn right and walk to the stove.'' 
This highlights the need for generating detailed, step-by-step action sequences for robotic agents to use in their execution processes.


\textit{Can LMs also learn grounded planning ability? How should we evaluate and improve LMs for grounded planning?}
In this paper, we address the question of whether LMs can also learn grounded planning abilities. To this end, we propose a study on the ability of language models for grounded planning for embodied tasks (\taskname{}). Our approach involves providing LMs with two inputs: a high-level task description and a realistic environment in the form of an object table. The output is a plan consisting of executable, step-by-step actions. We formulate \taskname{} as a language generation task and focus on encoder-decoder language models such as BART~\cite{Lewis2020BARTDS}.

In order to establish a dataset and evaluation protocol for \taskname{},
we leveraged the ALFRED data by developing a suite of data conversion programs.
They extract the object information from the environment and format it into data tables, 
thereby enabling models to access observations from realistic scenarios. Additionally, we formulated a new evaluation metric, referred to as \metricname{}, that is more appropriate for the task than existing ones for text generation.
As regards the methodology of \taskname{}, we suggest flattening an object table into a sequence of tokens and appending it to the task description as input to the model. 
The base LMs are then fine-tuned with these seq2seq data to learn to generate plans. Furthermore, we propose a simple yet effective decoding strategy that iteratively generates subsequent steps by incorporating the previous generation into the input.
Our empirical results and analysis indicate that incorporating object tables into inputs and the proposed iterative decoding strategies are both crucial for enhancing the performance of language models in \taskname{}.

To summarize, our main contributions are:
\begin{itemize}
    \item  
    {The task of \taskname{}}: To the best of our knowledge, this is one of the first studies to investigate the ability of LMs for embodied planning in realistic environments. \taskname{} is crucial for advancing the grounded generalization of large LMs and bridging the gap between NLP and embodied intelligence. (Sec.~\ref{sec:problem})
    \item  
    A comprehensive evaluation protocol: We put significant effort to convert the ALFRED and AI2THOR data into data tables to support the evaluation of \taskname{}. We also created a new evaluation metric, \metricname{}, to effectively assess the plans generated by the LMs.
    \item  
    Improving LMs for \taskname{}: We present two simple but effective components for enhancing the grounded planning ability of LMs - flattening object tables and an iterative decoding strategy. Our experiments show that these components lead to notable performance gains. (Sec.~\ref{sec:methods}) 
    Also, through extensive experimentation and in-depth analysis, we have gained a deeper understanding of the behavior of LMs for \taskname{} and present a series of non-trivial findings in our study.
\end{itemize}

\section{Problem Formulation}
\label{sec:problem}

\begin{figure*}[t]
\centering
\includegraphics[width=2\columnwidth]{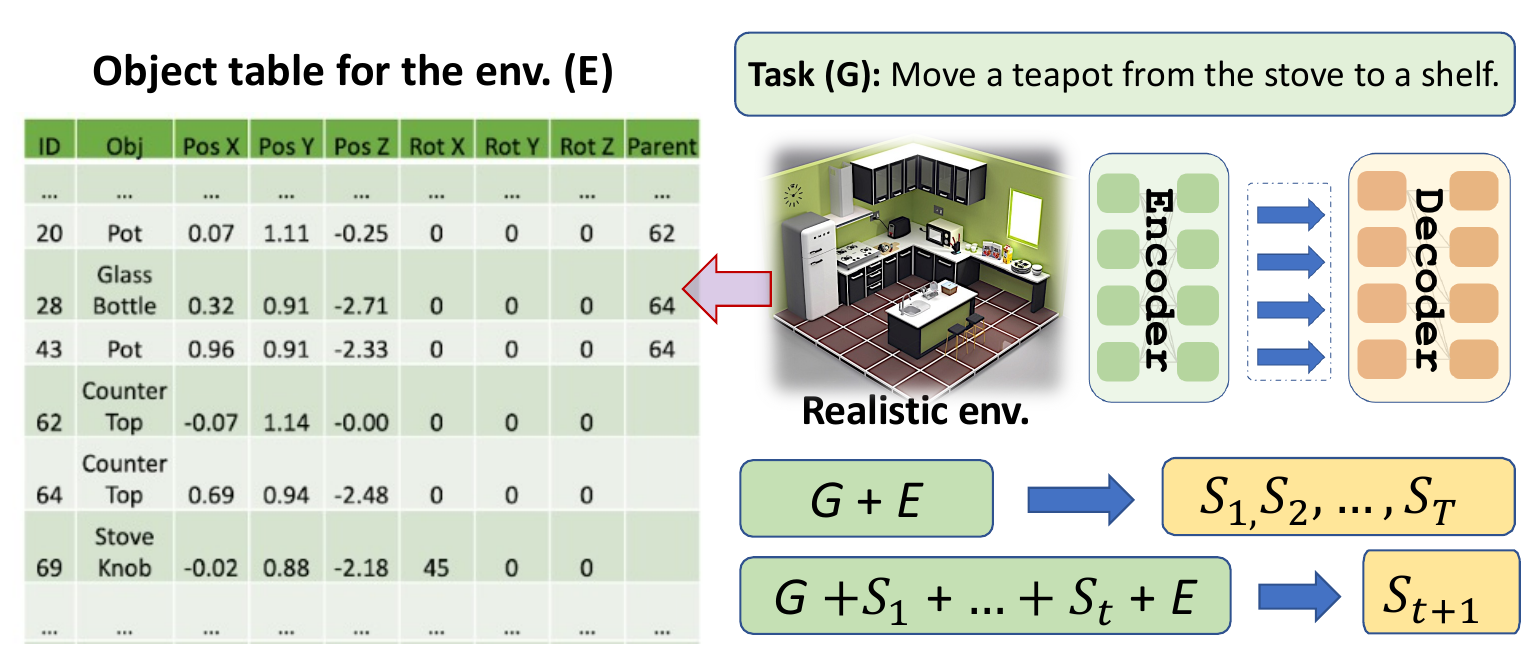}
\caption{The overall workflow of the proposed methods. First, we extract the object table from the realistic environment. Then we flatten the table into a sequence of tokens $E$ (Sec.~\ref{ssec:env_encoding}). 
We provide two learning methods for generating plans: 1) generate the whole plan $S_1,S_2,\cdots,S_T$ and 2) iteratively decode the $S_{t+1}$ (Sec.~\ref{ssec:iter_decoding}).}
\label{fig:pipeline}
\end{figure*}
Here we present the background knowledge, the problem formulation and the data sources for G-PlanET.

\subsection{Background Knowledge}
\label{ssec:background}
\paragraph{Embodied tasks.} The ALFRED benchmark~\cite{ALFRED20} is among the first benchmarks focusing on embodied tasks in realistic environments, although most of the examples are \textit{household} tasks. 
It aims to test the ability of agents to execute embodied tasks in real-world scenarios.
Specifically, the agents need to understand language-based instructions and output a sequence of actions to interact with an engine named AI2-THOR~\cite{ai2thor}, such that the given tasks can be achieved. 

\paragraph{Language instructions.}
Language instructions play an important role in the ALFRED benchmark.
The embodied tasks are annotated with a high-level goal and a low-level plan (\ie{} a sequence of executable actions for robots) in natural language, which are both inputs to the agents.
The agents need to understand such language instructions and parse them into action templates. 
Note that the agents do not need to \textit{plan} for the task, as they already have the step-by-step instructions to follow.

\paragraph{Task planning.}
Prior works show that large pre-trained language models (LMs) such as GPT-3~\cite{gpt3} can generate general procedures for completing a task. 
However, such plans are not aligned with the particular environment in which we are interested. 
This is because these methods never encode the environment as part of the inputs to LMs for grounding the plans to the given environment. 
Therefore, such non-grounded plans are hardly useful in guiding agents to work in real-world situations.

\subsection{G-PlanET with LMs}
\label{ssec:gplanet}
As discussed in Sec.~\ref{ssec:background}, the ALFRED benchmark does not explicitly test the \textit{planning} ability, while prior works on planning with LMs have not considered \textit{grounding} to a specific environment. 
In this work, we focus on evaluating and improving the ability to generate \textit{grounded plans} for \textit{embodied tasks} with LMs, which we dub as \taskname{}. 
It has been an underexplored open problem for both the robotics and NLP communities. 

\paragraph{Task formulation.}
The task we aim to study in this paper is essentially a language generation problem. Specifically, the input is two-fold: 1) a high-level goal $G$ and 2) a specific environment $E$ that the agents need to ground to. The expected output is a sequence of actionable plans $S=\{S_1, S_2, \cdots\}$ to solve the given goal in the specific environment step-by-step. 
The goal $G$ and the plan $S$ are in the form of natural language, while the environment $E$ can be viewed as a data table consisting of the object information in a room. 
Figure~\ref{fig:pipeline} shows an illustrative example and we will discuss more details in Section~\ref{ssec:env_encoding}.

\subsection{Data for \taskname{}}
\label{ssec:data_source}
To build a large-scale dataset for studying the \taskname{} task, 
we re-use the goals and the plans of ALFRED and extract object information from AI2THOR for the aligned environment.
The ALFRED dataset uses the AI2THOR engine to provide an interactive environment for agents with an egocentric vision to perform actions.
However, the dataset does not contain explicit data about objects in the environment (\eg{} the coordination, rotation, and spatial relationship with each other). 

We develop a suite of conversion programs for using AI2THOR to re-purpose the ALFRED benchmark for evaluating the methods shown in Section~\ref{sec:methods}.
We managed to get a structured data table to describe the environment of each task in the ALFRED dataset. 
We explore the AI2THOR engine and write conversion programs such that we can get full observations of all objects: properties (movable, openable, etc.), positions (3D coordinates \& rotation),  sizes, and spatial relationships (\eg{} object A is on the top of object B). 
We believe our variant of the ALFRED data will be a great resource for the community to study \taskname{} and future directions in grounded reasoning.


\section{Methods}
\label{sec:methods}
Herein, we introduce the methods that we adopt or propose to address the \taskname{} problem. 
First of all, we present the base language models that are encoder-decoder architectures. 
Then, we show in detail how we encode the environment data and integrate them with the seq2seq learning frameworks. 
Finally, we propose an interactive decoding strategy that significantly improves performance. 

\subsection{Base Language Models} 
Pretrained encoder-decoder language models, such as BART~\cite{Lewis2020BARTDS} and T5~\cite{Raffel2020ExploringTL}, have achieved promising performance in many well-known language generation tasks such as summarization and question answering. 
They also show great potential for general commonsense reasoning tasks such as CommonsenseQA~\cite{Talmor2019CommonsenseQAAQ}, suggesting that these large LMs have common sense to some extent. 
As the \taskname{} can be also viewed as a text generation problem, we use these LMs as the backbone for developing further planning methods, hoping that their common sense can be grounded in real-world situations for embodied tasks.  

\paragraph{Vanilla baseline methods.}
As shown in many papers, BART and T5, when sizes are similar, show comparable performance in many generation tasks. 
Thus, we use BART-base and BART-large as two selected LMs for evaluation.
The simplest and most straightforward baseline method of using such LMs to solve \taskname{} is to ignore the environment and only use the goal as the sole input.
Then, we fine-tune the base LMs with the training data and expect they can directly output the whole plan as a single sequence of tokens (including special separator tokens).
This simple method does not allow the LMs to perceive the environment, although training from the large-scale data can still teach the LMs some general strategies for planning.
Therefore, we see this as an important baseline method to analyze. 

\subsection{Encoding Realistic Environments}
\label{ssec:env_encoding}
To enable the LMs to perceive an environment, 
we need to encode the object tables described in Sec.~\ref{ssec:gplanet}.
Following prior works in table-based NLP tasks~\cite{Chen2020TabFact,Liu2021TAPEXTP}, 
we flatten a table into token sequences row by row, thus creating a linearized version of an object table. 
Then, we append the flattened table after the goal to form a complete input sequence. 
Thus, the input side of the encoder-decoder finally has the environment information for generating a grounded plan.

Considering the max sequence limit, we only choose to encode objects by their \textbf{type}, \textbf{position}, \textbf{rotation}, and the \textbf{receptacle} parent.
The object type does not only tell what an object is but also implies commonsense affordance (\eg{} a microwave can heat up something, a knife can slice something) which is very important for planning.
The position information is essential for agents to navigate and find objects, thus playing an important part in planning.
The rotation is also useful for some objects that can only be used with a certain orientation (\eg{} a refrigerator can only be opened when the agent is in front of it). 
The receptacle of an object and itself has a close spatial connection (\eg{} a pen is on a desk; an apple is in a fridge).
Every object has a unique identifier such that objects of the same type can be referred to precisely when they are receptacles of others. 
In addition, the agent is represented as a special object.






\subsection{Iterative Decoding Strategy}
\label{ssec:iter_decoding}
Adding the flattened table of object information to the input sequences indeed improves the LMs in terms of their perception of the realistic environments, which forms the foundation of grounded planning. 
However, the thinking process is still limited by the conventional seq2seq learning framework, which assumes LMs should output a complete plan by a single pass of decoding.
We argue that a thoughtful planning process should carefully handle the coherence of each step, otherwise errors accumulate and cause a failed plan. 

Therefore, we propose a simple yet effective decoding strategy that learns to iteratively generate a plan step by step.  
Specifically, we append previously generated steps until the current step $t$ to the input sequence (\ie{} \texttt{Input} = $[G + S_1 + \cdots + S_t (+ E)]$) for generating the next step (\ie{} \texttt{Output} = $S_{t+1}$).
This iterative decoding process will end until the LM generates the special token 
\texttt{END}. 
In the training stage, we use the ground-truth references for $S_{\le t}$; in the inference stage, we do not have such references, so we use the model predictions as $S_{\le t}$.

Notably, in contrast to the conventional seq2seq learning process, the iterative decoding strategy needs to run the encoder-decoder model $N+1$ times to generate a plan with $N$ steps. 
The additional computation cost for re-encoding is worthy. 
Imagine when we humans are planning a task in a room. 
It is natural for us to come up with the plans step by step, and it is very likely that the most useful information to generate different steps is about different objects.
Therefore, a temporally dynamic attention mechanism is favorable in planning with LMs.
Our iterative decoding strategy encourages the encoder-decoder architectures to learn such ability.   


\subsection{Other Methods}
\paragraph{Pretrained table encoders.}
Since we use environmental information in a tabular format and BART has not been pre-trained in the tabular form of input, BART may not be able to use this part of information well. Therefore, we employ \textsc{TaPEx} \cite{Liu2021TAPEXTP}, the state-of-the-art pre-trained language model on tabular data.
Using SQL execution as the only pre-training task, \textsc{TaPEx} achieves better tabular reasoning capability than BART, and thus we expect \textsc{TaPEx} can make full use of the environmental information represented by the table in our task.
\paragraph{In-context few-shot learning with GPT-J.}
Finally, to explore whether large-scale language models can master the task with few-shot examples, we also experimented with few-shot performance on a larger language model GPT-J 6B.

\section{Evaluation}
\label{sec:evaluation}



How do we evaluate a method for \taskname{}? 
Due to the novelty of the problem setup, it is challenging to evaluate and analyze the methods. 
In this section, we present a general evaluation protocol and a complementary metric to measure the quality of generated plans. 
We report the main experimental results with the proposed evaluation protocol.
We leave the analysis in Sec.~\ref{sec:analysis}.

\subsection{Metrics}
\label{ssec:metrics}

\paragraph{Step-wise evaluation.}
Conventional evaluation metrics such as BLEU~\cite{Papineni2002BleuAM} and ROUGE~\cite{Lin2004ROUGEAP} measure the similarity between  generated text and truth references as a whole, which is suitable for translation and summarization.
However, the output text of planning tasks such as our~\taskname{} is highly structured. 
A plan naturally can be split into a sequence of step-by-step actions. 
Using the conventional way to evaluate plans inevitably breaks such internal structures and will lead to inaccurate measurement. 
For example, if the first step of the generated plan is the same as the last step of the reference plan, the conventional evaluation will still assign a high score to such a generated plan, even though it is not useful at all. 
Therefore, we argue that it is much more reasonable to evaluate the similarity of a pair of plans step by step. 
Specifically, we first align the generations and the truths and compute the scores of every step\footnote{{The ALFRED authors ensure that the references consist of atomic action steps and all references share the same length. Therefore, we consider the length of truth plans as the standard:  when the generated plan has more steps than the truth plans, we cut off them; when the generation has fewer steps than the references, we duplicate the last step to make them even for step-wise evaluation. } } by multiple metrics. Then, we aggregate the final score by taking the average of all steps. 
We also consider other temporal weighting aggregation for more analysis in Sec.~\ref{sec:analysis}.

\begin{table*}[th]
\centering
\begin{tabular}{c|ccc|ccc}
\toprule
\multicolumn{1}{r}{\textbf{Data Split} $\rightarrow$}                                             & \multicolumn{3}{c}{Unseen Room Layouts} & \multicolumn{3}{c}{Seen Room Layouts} \\ \midrule
\multicolumn{1}{r}{\textbf{Methods} $\downarrow$ \textbf{Metrics} $\rightarrow$} & CIDEr   & SPICE   & KAS    & CIDEr  & SPICE  & KAS    \\ \midrule
BART-base  (vanilla)                                                                                               & 0.9417  & 0.1378  & 0.2455 & 0.8231 & 0.1277 & 0.2197 \\
BART-large (vanilla)                                                                                               & 1.4632  & 0.3168  & 0.4069 & 1.4414 & 0.3161 & 0.3900 \\
GPT-J-6B                                                                                                           & 1.1968  & 0.2655  & 0.3622 & 1.1047 & 0.2509 & 0.3370 \\  \midrule
BART-base w/table                                                                                                  & 1.6706  & 0.3692  & 0.4584 & 1.6230 & 0.3595 & 0.4339 \\
BART-large w/table                                                                                                 & 1.6630  & 0.3491  & 0.4411 & 1.5865 & 0.3393 & 0.4204 \\
BART-large (\textsc{TaPEx})                                                                       & {2.8824}  & 0.5054  & 0.6373 & 2.7432 & 0.4944 & 0.6045 \\ \midrule
BART-base w/table + iterative decoding                                                                             & \textbf{2.9147}  & 0.5107  & 0.6334 & 2.8582 & \textbf{0.5118} & 0.6124 \\
BART-large w/table + iterative decoding                                                                            & 2.8580  & 0.5194  & \textbf{0.6518} & \textbf{2.8799} & 0.5096 & \textbf{0.6326} \\
BART-large (\textsc{TaPEx})  + iterative decoding                                                  & 2.8440  & \textbf{0.5210}  & 0.6313 & 2.6959 & 0.5036 & 0.6074\\ \bottomrule
\end{tabular}
\caption{{Experimental results for the~\taskname{} by different base LMs. 
The methods are grouped by model types and whether encoding the environment; by decoding strategies.}}\label{tab:main_result}
\end{table*}

\paragraph{Measuring grounded plans.}
It is a unique challenge for evaluating~\taskname{} to consider the grounding nature of plans. 
Metrics, such as BLEU, METEOR, and ROUGE, do not give a suitable penalty when a plan is similar to the reference in terms of word usage, yet leading to totally different states in an interactive environment for embodied tasks. 
For example, it is only a one-word difference between ``turn to the \textit{left}'' vs ``turn to the \textit{right}'', but the agents that faithfully follow these instructions can arrive at very different places.

The LM-based metrics, \eg{} BERTScore~\cite{Zhang2020BERTScoreET}, are not suitable either because the neural embeddings of ``left'' and ``right'' are also very similar. 
Plus, the grounded plans for \taskname{} are object-centric in a context and very similar to the captions of a sequence of events by visual perception, 
for which these metrics are not specifically designed.
Considering these limitations, we use two typical metrics that are widely used for captions and devise a new metric for complementary measurement.

The first two metrics are \textbf{CIDEr}~\cite{Vedantam2015CIDErCI} and \textbf{SPICE}~\cite{Anderson2016SPICESP}, which are both widely used for tasks where the outputs are highly contextualized and describe natural scenarios in everyday life, \eg{} VaTex~\cite{Wang2019VaTeXAL} and CommonGen~\cite{Lin2020CommonGenAC}.
In particular,
SPICE parses both the generation and references to scene graphs, a graph-based semantic representation. 
Then, it calculates the edge-based F1 score to measure the similarity between each step.
Note that SPICE computation has a special focus on the propositions. 
This is particularly favorable for evaluating ~\taskname{} since there are many actions in the grounded plans, where propositions can be seen as atomic units for evaluation.



\paragraph{KeyActionScore (\metricname{}).} 
Inspired by SPICE, a step in a plan can be deconstructed into several propositions that are represented as edges. 
However, not all propositions in SPICE are necessarily important in evaluating plans for ~\taskname{}.
Not to mention that SPICE relies on an external parser that is expensive to run yet sometimes contains noisy outputs. 
Also, most of the truth plans in the ALFRED annotations are overly specific, and it is not necessary for a plan to cover all details. 
Therefore, we devise a metric that focuses on the key actions of the generated plans and checks if they are part of references, named \textit{Key Action Score} (\textbf{KAS}). 

Specifically, we extract a set of key action phrases from each step in the generated plan $\hat{S_i}$ and the truth reference $S_i$ respectively.
We denote this two sets as $\hat{S_i}=\{\hat{a}_1, \hat{a}_2, \cdots\}$ and ${S}_i=\{a_1, a_2, \cdots\}$.
Then, we check how many action phrases in $\hat{S_i}$ are covered by the truth set ${S_i}$, the precision then becomes the KAS score for the $i$-th step in the plan.
To increase the matching quality, we curate a set of rules and a dictionary to map the actions that share the same behaviors. 
For example, ``\textit{turn} to the \textit{left}'' and ``turn left'' are counted as a single match; ``\textit{go} straight'' and ``\textit{walk} straight'' can be matched too.
In addition, we break the compound nouns such that we allow partial scores to match for a smoother scoring (\eg{} ``xxx on the \textit{table}'' vs ``xxx on the \textit{coffee table}'').
Simply put,
the KAS metric looks at the key actions extracted from the plans and checks if these important elements can be (fuzzy) matched to count as a valid step.




\subsection{Experimental Setup}

\paragraph{Data statistics.}
Table ~\ref{tab:data_stat} shows some statistics of our dataset that we described in Sec.~\ref{ssec:gplanet}. 
We follow the data split in ALFRED to split the train, valid, and test dataset.
The data split is based on whether the room layout has been seen in the training tasks.
It is usually easier for robotic agents to map instructions to low-level actions in seen rooms than in unseen rooms.
However, for the planning ability that we want to study with~\taskname{} in this paper, the two splits do not differ very much. 
We keep using this split to make the results consistent and convenient for people who want to connect our results with the ALFRED results. 

\begin{table}[!t]
    \centering
\scalebox{0.88}{
    \begin{tabular}{r|c|c|c|c|c@{}}
    \toprule 
        \textbf{split} $\rightarrow$        & \textbf{train} & \multicolumn{2}{c|}{\textbf{valid}} & \multicolumn{2}{c}{\textbf{test}} \\
\textbf{aspect} $\downarrow$ & -     & seen        & unseen      & seen       & unseen      \\\midrule
\# tasks     & 21,025 & 820         & 821         & 705        & 694         \\
\midrule
avg. $|G|$      & 9.26  & 9.32        & 9.26        & 10.3       & 9.95       \\
avg. \# $O$ & 73.71 & 74.21       & 77.91       & 75.31      & 73.9        \\
\midrule
avg. \# $T$   & 6.72  & 6.79        & 6.26        & 6.95       & 6.63        \\
avg. $|S_i|$    & 11.24 & 11.13       & 11.49       & 9.84       & 10.19       \\
 \bottomrule
\end{tabular}
}
    \caption{The avg. $|G|$ means the average length of goal and the avg. $|S_i|$ means the average length of each step. The avg. \# $O$ is the average number of objects in each room and the avg. \# $T$ is the average number of steps.}
    \label{tab:data_stat}
\end{table}

\paragraph{Implementation details.}
In single-pass decoding, 
we format the output sequences as follows:
``$\texttt{Step~1:} [S_1]\ | \ \texttt{Step~2:}\ [S_2]\ | \ \cdots \ |\ \texttt{END}$''. 
When appending the flattened table of objects, 
we format input with 
``$[G]~\texttt{Env:}~[\texttt{row 1}]~[\texttt{SEP}]~[\texttt{row 2}]~\cdots$'',
where the $[\texttt{row i}]$ is a sequence of the $i$-th object including its id, type, coordinates, rotation, parent receptacles, etc.
Due to the page limit, we leave the details of the data, methods, and hyper-parameters in the Appendix that are linked to our project website.



\subsection{Main Results}
\label{ssec:main_results}

We report the main results in Table \ref{tab:main_result}, and leave the deeper analysis in the next section.
To sum up, we find that encoding the object table as part of the inputs will significantly improve the performance, and pre-training on other table-related tasks can benefit~\taskname{} a lot.
The iterative decoding strategy is also an important component that can further improve the results to some extent.

\paragraph{Case Study of Table Effect}
Although we have added environment information $E$ to the input, it is still a problem whether the model effectively uses this information. To verify this, we present a case study here. 
In a number of instances, we have demonstrated that the introduction of environmental information can be helpful. 
Here is one example: 
\begin{itemize}
    \item \textbf{truth:} Close the laptop that is on the \emph{table}.
    \item \textbf{vanilla:} Close the laptop and pick it up from the \emph{bed}
    \item \textbf{w/ table:} Pick up the laptop on the coffee \emph{table}. 
\end{itemize}
As shown in the example, the model successfully identified the location of a laptop with the help of the object table. 

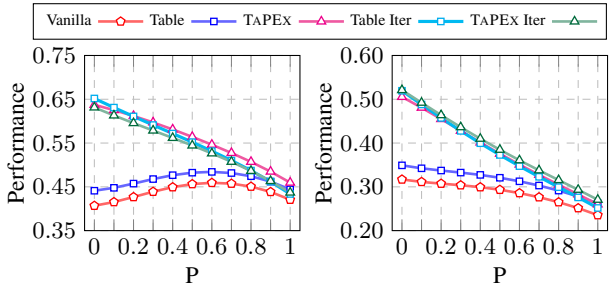
\begin{figure}[t!]
\centering
\begin{tikzpicture}
\small{
\begin{axis}[
at={(0,0)},
width=.25\textwidth, height=.22\textwidth ,
xtick={0, 1, ..., 10},
xticklabels={
 {$0$}, {$ $}, {$0.2$}, {$ $}, {$0.4$},
 {$ $}, {$0.6$}, {$ $}, {$0.8$}, {$ $},{$1$}
},
ytick={0.35,0.45,...,0.75},
yticklabels={{$0.35$}, {$0.45$}, {$0.55$}, {$0.65$}, {$0.75$}},
grid style=dashed,
ylabel={Performance},
xlabel={{P}},
xlabel style={yshift=0.5em},
ylabel style={align=center,yshift=-1em},
y tick style={opacity=0},
ymajorgrids=true,
xmajorgrids=true,
tick align=inside,
legend pos=outer north east,
yticklabel style={/pgf/number format/precision=1,/pgf/number format/fixed zerofill},
legend columns=5,
legend style={yshift=2.2em,xshift=-11.5em,legend cell align=left,legend plot pos=right},
xmin=-0.5,
xmax=10.5,
ymin=0.35,
ymax=0.75]
    \addplot[
        red!60,mark=pentagon*,mark size=1.7pt,thick,line width=1pt,mark options={fill=white,draw=red,line width=0.5pt}
        ]
        coordinates {
        (0, 0.40685786346056446)
        (1, 0.41523220586792153)
        (2, 0.42671737113615565)
        (3, 0.43883711798512176)
        (4, 0.4492341277170484)
        (5, 0.4563061817931629)
        (6, 0.45916093375245354)
        (7, 0.45733663176010414)
        (8, 0.4505351926777318)
        (9, 0.4384400699718185)
        (10, 0.4206321243910976)
        };
      \addplot[
        blue!60,mark=square*,mark size=1.2pt,thick,line width=1pt,mark options={fill=white,draw=blue,line width=0.5pt}
        ]
        coordinates {
        (0, 0.4411072536416398)
        (1, 0.4477302812304057)
        (2, 0.4573874754646058)
        (3, 0.4677515345257907)
        (4, 0.47658510323240705)
        (5, 0.48234079287036347)
        (6, 0.48413878897234036)
        (7, 0.48152747936186824)
        (8, 0.4742257809012899)
        (9, 0.46191844563783574)
        (10, 0.44413941671215884)
        };
        \addplot[
        magenta!80,mark=triangle*,mark size=1.9pt,thick,line width=1pt,mark options={fill=white,draw=magenta,line width=0.5pt}
        ]
        coordinates {
        (0, 0.63730132336551)
        (1, 0.6252446446223785)
        (2, 0.6118969467454409)
        (3, 0.5970934713178261)
        (4, 0.5810800806423883)
        (5, 0.5641119923532867)
        (6, 0.5462523453303506)
        (7, 0.5273563527073546)
        (8, 0.5070698744008714)
        (9, 0.4847928836386209)
        (10, 0.4596301617846205)
        };
        \addplot[
        cyan!80,mark=square*,mark size=1.2pt,line width=1.2pt,mark options={fill=white,draw=cyan,line width=0.5pt,solid}
        ]
        coordinates {
        (0, 0.6518061268982771)
        (1, 0.6310081972601632)
        (2, 0.6108115629179518)
        (3, 0.5911654441811401)
        (4, 0.571708009320217)
        (5, 0.5519964014174)
        (6, 0.5316195792573555)
        (7, 0.5101765666137057)
        (8, 0.48719155331517505)
        (9, 0.4620098568245687)
        (10, 0.43370671383475456)
        };
        \addplot[
        bronze!50,mark=triangle*,mark size=1.9pt,mark options={fill=white,draw=bronze,line width=0.5pt,solid},line width=1pt
        ]
        coordinates {
        (0, 0.6313238353241837)
        (1, 0.613039458142605)
        (2, 0.5955435232449245)
        (3, 0.5785869678628588)
        (4, 0.5617022894383822)
        (5, 0.5444420805282706)
        (6, 0.5264132412521253)
        (7, 0.5072224731768339)
        (8, 0.48639465276751453)
        (9, 0.4632746231858139)
        (10, 0.43693735285625257)
        };
        \legend{\tiny{Vanilla},\tiny{Table},\tiny{\textsc{TaPEx}},\tiny{Table Iter},\tiny{\textsc{TaPEx} Iter}}
\end{axis}
\begin{axis}[
at={(.23\textwidth,0)},
width=.25\textwidth, height=.22\textwidth ,
xtick={0, 1, ..., 10},
xticklabels={
 {$0$}, {$ $}, {$0.2$}, {$ $}, {$0.4$},
 {$ $}, {$0.6$}, {$ $}, {$0.8$}, {$ $},{$1$}
},
ytick={0.20,0.30,...,0.60},
yticklabels={{$0.20$}, {$0.30$}, {$0.40$}, {$0.50$}, {$0.60$}},
grid style=dashed,
ylabel={Performance},
xlabel={{P}},
xlabel style={yshift=0.5em},
ylabel style={align=center,yshift=-1em},
y tick style={opacity=0},
ymajorgrids=true,
xmajorgrids=true,
tick align=inside,
legend pos=outer north east,
yticklabel style={/pgf/number format/precision=1,/pgf/number format/fixed zerofill},
xmin=-0.5,
xmax=10.5,
ymin=0.20,
ymax=0.60]
    \addplot[
        red!60,mark=pentagon*,mark size=1.7pt,thick,line width=1pt,mark options={fill=white,draw=red,line width=0.5pt}
        ]
        coordinates {
        (0, 0.31684431192253737)
        (1, 0.31127295864973853)
        (2, 0.30730654717179157)
        (3, 0.3036738615393242)
        (4, 0.2992416867239351)
        (5, 0.29335238882439907)
        (6, 0.28573847389556084)
        (7, 0.276303063943734)
        (8, 0.26493372311589475)
        (9, 0.25140080465709486)
        (10, 0.23534790678798137)
        };
      \addplot[
        blue!60,mark=square*,mark size=1.2pt,thick,line width=1pt,mark options={fill=white,draw=blue,line width=0.5pt}
        ]
        coordinates {
        (0, 0.34905618386844584)
        (1, 0.3423302540918126)
        (2, 0.3372216737116396)
        (3, 0.3325517586078169)
        (4, 0.32726427661208213)
        (5, 0.3207197555083438)
        (6, 0.3126351596203602)
        (7, 0.3029063009038285)
        (8, 0.2914268381611666)
        (9, 0.27796327133939286)
        (10, 0.26211806289943684)
        };
        \addplot[
        magenta!80,mark=triangle*,mark size=1.9pt,thick,line width=1pt,mark options={fill=white,draw=magenta,line width=0.5pt}
        ]
        coordinates {
        (0, 0.5054096131035745)
        (1, 0.48080073713441834)
        (2, 0.45481013789425156)
        (3, 0.4280658395134587)
        (4, 0.4014278522284052)
        (5, 0.37557793578532606)
        (6, 0.3508742223154198)
        (7, 0.32736594736113894)
        (8, 0.30480453756203135)
        (9, 0.28262567814911094)
        (10, 0.2599375470221303)
        };
        \addplot[
        cyan!80,mark=square*,mark size=1.2pt,line width=1.2pt,mark options={fill=white,draw=cyan,line width=0.5pt,solid}
        ]
        coordinates {
        (0, 0.5194285728933805)
        (1, 0.48814104642248246)
        (2, 0.45736405379712336)
        (3, 0.4277606630293626)
        (4, 0.39956854815980386)
        (5, 0.372778992163757)
        (6, 0.34728492801556404)
        (7, 0.3228919286213619)
        (8, 0.2992559726595239)
        (9, 0.27580931950553117)
        (10, 0.2517162978194509)
        };
        \addplot[
        bronze!50,mark=triangle*,mark size=1.9pt,mark options={fill=white,draw=bronze,line width=0.5pt,solid},line width=1pt
        ]
        coordinates {
        (0, 0.5209648071334182)
        (1, 0.4921390463810853)
        (2, 0.46371599953965337)
        (3, 0.43621823556382405)
        (4, 0.409859013588879)
        (5, 0.3846932195206426)
        (6, 0.3606862708289208)
        (7, 0.3376979006340092)
        (8, 0.315422690537387)
        (9, 0.2933169355692448)
        (10, 0.2705489197507247)
        };
\end{axis}
}
\end{tikzpicture}
\caption{The step-wise reweighting results of \metricname{} (\textbf{Left}) and SPICE (\textbf{Right}).The x-axis indicates the parameter p in the geometric distribution and also the importance of the preceding step, and the y-axis indicates the weighted result of each step. A larger coefficient means that the previous step is more important.}
\label{fig:step_spice}
\end{figure}

\paragraph{Effect of model sizes.}
Table \ref{tab:main_result} shows that small models can perform as well or even better than large models in some cases. This is mainly due to the following reasons. 1) The sentences in plans are relatively simpler than other NLG tasks, with a smaller vocabulary and shorter length. This leaves the power of large models in terms of generation unexpressed, 2) \taskname{} is a task to examine the ability to plan rather than write. Whether this ability changes with model size remains to be explored. 3) For scenarios with the table, the form of the task is not the same as the traditional generation task, so the training phase will have a greater impact. 
Models with fewer parameters are more sufficiently tuned with limited data.

\section{Analysis}
\label{sec:analysis}


In this section, we deeply analyze the performance of the methods in Table~\ref{tab:main_result} from multiple aspects and provide non-trivial findings that can help future research. For a fair comparison, all analytical experiments were performed in the BART-large model on the unseen split of the test data. 

\subsection{Temporal Re-weighting of Scores}
When we computed the overall score of a plan with a metric,  
we use the average score to aggregate the score for each step.
However, in a realistic environment, there are causality constraints for an agent to complete the steps -- \ie{} some tasks can only be done when their prerequisite steps are finished.
For example, only when the agent arrives at the microwave can it heat the bread in its hands. 

Therefore, the earlier steps in a plan should  be of higher importance, while our previous evaluation is based on a uniform distribution of the weights across steps.
To this end, we adopt \emph{geometric distribution} to re-weight the step-wise importance for weighted aggregation. 
The geometric distribution can be used to model the number of failures before the first success in repeated mutually independent Bernoulli trials, each with a probability of success $p$.
\begin{align*}
    f(x) = p(1-p)^x \qquad (0 < p < 1)
\end{align*}

\pgfplotstableread[row sep=\\,col sep=&]{
    Scale & vanilla & table & tapex & table_iter & tapex_iter \\
    0  & 69 & 72 & 23 & 8 & 8 \\
    1 & 81 & 78 & 48  & 19 & 21 \\
    2  & 149 & 98 & 100 & 114 & 126 \\
    3  & 246 & 281  & 498 & 518 & 517 \\
    4 & 148 & 165 & 25 & 35 & 22 \\
    5 & 0 & 0 & 0 & 0 & 0 \\
}\datatable

\begin{figure}[t!]
    \small
    \centering
    \begin{tikzpicture}
        \begin{axis}[
                ybar,
                enlarge x limits=0.005,
                ybar interval=0.75,
                width=0.45\textwidth,
                height=.2\textwidth,
                bar width=0.8em,
                legend style={at={(0.5,1.3)},
                anchor=north,legend columns=-1,font=\scriptsize},
                ylabel style={align=center,yshift=-1em},
                ytick={0,100,...,550},
                y tick style={opacity=0},
                xtick={0,1,2,3,4,5},
                xticklabels={
             {$(-\infty,-5]$}, {$(-5,-3]$}, {$(-3,-1]$}, {$0$}, {$(0,+\infty)$}, {}
                },
                ymajorgrids=true,
                xmajorgrids=true,
                grid style=dashed,
                tick align=inside,
                xmax=5,
                xmin=0,
                ymin=0,ymax=550,
                ylabel={Count},
                xlabel={Step Range}
            ]

            \addplot +[fill=red!25,draw=red] table[x=Scale,y=vanilla]{\datatable};
            \addplot +[fill=blue!25,draw=blue] table[x=Scale,y=table]{\datatable};
            \addplot +[fill=magenta!25,draw=magenta] table[x=Scale,y=tapex]{\datatable};
            \addplot +[fill=cyan!25,draw=cyan] table[x=Scale,y=table_iter]{\datatable};
            \addplot +[fill=bronze!25,draw=bronze] table[x=Scale,y=tapex_iter]{\datatable};
            
            \legend{Vanilla, Table, \textsc{TaPEx}, Table Iter, \textsc{TaPEx} Iter}
        \end{axis}
    \end{tikzpicture}
    \caption{The result of error statistics for \# of step.}
\label{fig4}

\end{figure}
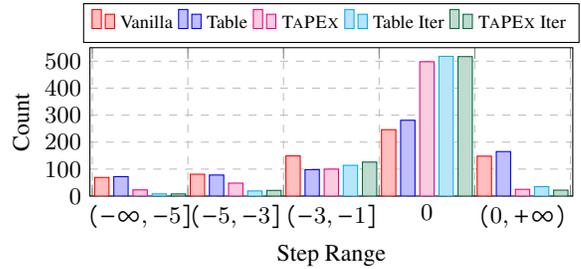


This suits our setup well because when the first step is incorrect, the whole task can hardly be completed and executed in a generated plan for ALFRED. 
The range of $p$ in the original setting of the geometric distribution is restricted to between $0$ and $1$. 
When $p=0$, each step has the same weight (uniform importance), which is exactly what we have done in Tab.~\ref{tab:main_result}. 
When $p=1$, the first step is the only thing we look at for evaluation, meaning that the other steps will be given zero weights for aggregation.

Figure \ref{fig:step_spice} shows the results on unseen subset which is more realistic. The performance of the iterative and non-iterative approaches is very close in the case of the first step. This is mainly because iterative methods are similar to non-iterative methods when generating the first step, and differ only after the second step. 
At the same time, it can be seen that there is an overall downward trend in performance as the focus moves to the early step. The main reason is that the later the subtask is, the closer it is to the high-level instruction. For example, if the task goal is to place the sponge in the sink, the final step must be to place the sponge in the sink. This feature makes the last step of subtask generation very simple, resulting in high performance. 
We also see that the performance of the non-iterative method rises and then falls in \metricname{}, and the change in a downward trend in SPICE. 
The main reason is an error in the number of steps in the non-iterative method, which will be explained next.

\subsection{Error Analysis on the Lengths of Plans}

We found a huge gap in the prediction of the number of task steps between iterative and non-iterative methods, which may be an important reason for the final performance difference. As shown in Figure \ref{fig4}, iterative methods have a higher probability of predicting the number of steps for the correct task, while non-iterative methods do underestimate the number of steps. In our evaluation framework, the missing follow-up steps of non-iterative methods are often generated by copying. This might be reason for the poor performance of non-iterative methods and the performance of non-iterative methods increases first in the reweight step process.

\subsection{Impact of Task Length on Performance}
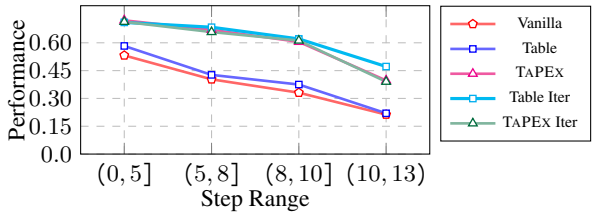
\begin{figure}[t!]
\hspace{-1em}
\centering
\begin{tikzpicture}
\small{
\begin{axis}[
at={(0,0)},
width=.35\textwidth, height=.2\textwidth ,
xtick={0, 1, 2, 3},
xticklabels={
 {$(0,5]$}, {$(5,8]$}, {$(8,10]$}, {$(10,13)$}
},
ytick={0,0.15,0.30,...,0.75},
yticklabels={{$0.0$}, {$0.15$}, {$0.30$}, {$0.45$}, {$0.60$}, {$0.75$}},
grid style=dashed,
ylabel={Performance},
xlabel={{Step Range}},
xlabel style={yshift=0.5em},
ylabel style={align=center,yshift=-1.2em},
y tick style={opacity=0},
ymajorgrids=true,
xmajorgrids=true,
tick align=inside,
legend pos=outer north east,
yticklabel style={/pgf/number format/precision=1,/pgf/number format/fixed zerofill},
xmin=-0.5,
xmax=3.5,
ymin=0.0,
ymax=0.8]
    \addplot[
        red!60,mark=pentagon*,mark size=1.7pt,thick,line width=1pt,mark options={fill=white,draw=red,line width=0.5pt}
        ]
        coordinates {
        (0, 0.532)
        (1, 0.403)
        (2, 0.331)
        (3, 0.213)
        };
      \addplot[
        blue!60,mark=square*,mark size=1.2pt,thick,line width=1pt,mark options={fill=white,draw=blue,line width=0.5pt}
        ]
        coordinates {
        (0, 0.583)
        (1, 0.427)
        (2, 0.375)
        (3, 0.220)
        };
        \addplot[
        magenta!80,mark=triangle*,mark size=1.9pt,thick,line width=1pt,mark options={fill=white,draw=magenta,line width=0.5pt}
        ]
        coordinates {
        (0, 0.721)
        (1, 0.670)
        (2, 0.605)
        (3, 0.397)
        };
        \addplot[
        cyan!80,mark=square*,mark size=1.2pt,line width=1.2pt,mark options={fill=white,draw=cyan,line width=0.5pt,solid}
        ]
        coordinates {
        (0, 0.710)
        (1, 0.684)
        (2, 0.621)
        (3, 0.472)
        };
        \addplot[
        bronze!50,mark=triangle*,mark size=1.9pt,mark options={fill=white,draw=bronze,line width=0.5pt,solid},line width=1pt
        ]
        coordinates {
        (0, 0.713)
        (1, 0.658)
        (2, 0.612)
        (3, 0.390)
        };
        \legend{\tiny{Vanilla},\tiny{Table},\tiny{\textsc{TaPEx}},\tiny{Table Iter},\tiny{\textsc{TaPEx} Iter}}
\end{axis}
}
\end{tikzpicture}
\caption{The result of \metricname{} of tasks with a different number of steps. Due to the large variance caused by the small number of samples of certain lengths, we use the statistics by dividing the intervals.}
\label{fig5}
\end{figure}

Although all the tasks in the dataset are part of daily life tasks, they differ in difficulty. A simple metric to evaluate the difficulty of a task is the number of steps they require. Figure \ref{fig5} illustrates the decrease in the quality of the generated steps as the number of task steps increases. The figure also reflects the relatively small difference in the performance of the different methods on shorter tasks. 
And the performance of all methods degrades rapidly on the longest tasks. The iterative approach has more significant performance benefits on longer tasks. This may be because this approach makes better use of the state changes due to intermediate steps and fixes some previous errors.

\section{Related Work}

\paragraph{Grounded commonsense reasoning.}
ALFWorld ~\cite{Shridhar2021ALFWorldAT} also uses LM to generate the next step in a text game which is based on ALFRED. SciWorld~\cite{scienceworld2022} designed a text game to find whether the LMs have learned to reason about commonsense.  SayCan~\cite{saycan2022arxiv} also uses LM to find the potential next step in the real world. Both these three works only expect to learn the next step in a text game. Our methods share similar motivation with decision transformer~\cite{Chen2021DecisionTR} and Behavior Cloning~\cite{Farag2018BehaviorCF}, but we work on very different applications.

\paragraph{Table-based NLP.}
Our work is closely related to two lines of tabular data usage in NLP: the approach to modeling tabular representations and the application of a table as an intermediate representation.
For the first line of work, there is rich literature focusing on modeling tabular representations, including TabNet \cite{Arik2021TabNetAI}, TAPAS \cite{herzig-etal-2020-tapas}, TaBERT \cite{yin-etal-2020-tabert} and \textsc{TaPEx} \cite{Liu2021TAPEXTP}.
We have explored the impact of state-of-the-art table representation models (e.g., \textsc{TaPEx}) on our task in experiments.
As for the second line of work, previous work has explored to use of tables in several downstream tasks, including visual question answering \cite{vqa2018nips}, code modeling \cite{pashakhanloo2022codetrek}, and numerical reasoning \cite{Pi2022ReasoningLP,yoran-etal-2022-turning}.
Different from them, our work is the first to explore the use of tabular representations in embodied tasks.

\paragraph{ALFRED Agents.}
Some previous research has been published on embodied tasks in realistic environments since the appearance of ALFRED. E.T. ~\cite{Pashevich2021EpisodicTF} first encoded the history with a transformer to solve compositional tasks and proved that pretraining and joint training with synthetic instructions can improve performance. FILM ~\cite{Min2021FILMFI} proposed an explicit spatial memory and a semantic search policy to provide a more effective representation for state tracking and guidance. LEBP ~\cite{Liu2022LEBPL}, the currently published SOTA method, generated a sequence of sub-steps by understanding the language instruction and used the predefined actual actions template to complete the sub-steps. We also try to use these methods to evaluate our generated low-level instructions. However, due to the limited importance of the low-level instructions, there is no gap with conspicuousness between our generated instructions and the ones in ALFRED.

\section{Conclusion}

In this work, we present the first investigation into grounded planning for embodied tasks using language models. The \taskname{} problem is of utmost significance for advancing the embodied intelligence of LMs and constitutes a critical step towards artificial general intelligence.
To evaluate the performance of encoder-decoder LMs in solving \taskname{}, we developed a benchmark as well as a specialized evaluation metric named \metricname{} to assess the quality of generated plans.
Furthermore, we propose two methods for improving LMs' ability in \taskname{} - flattening object tables and an iterative decoding strategy. Our experiments and analyses demonstrate their effectiveness and yield non-trivial findings.
This study is expected to encourage further research into \taskname{} and pave the way for integrating LMs and embodied tasks in realistic environments.

The main limitations of this work on the new task~\taskname{} are as follows:
\begin{itemize}
    \item \textbf{Evaluation}: Although we have adopted and devised automatic metrics for evaluating methods for~\taskname{}, there is not yet a straightforward way for us to test the ultimate success rates of such plans (if they were executed by oracle agents). We tried using state-of-the-art ALFRED agents such as FILM~\cite{Min2021FILMFI}, but they did not show obvious differences using step-by-step instructions (even if using the oracle version). We believe more human evaluation will help us further refine the metrics, which can be very expensive though. It is because human annotators much play with the 3D engine while following these instructions, in order to assess the quality of such plans. 
    \item \textbf{Methods:} Flattening object tables into sequences of tokens row by row is straightforward but might not be optimal. The number of objects can be huge for a complicated room. How can we narrow down the important objects at each step? We argue that a more advanced version of attention modules for dynamic table encoding is needed. We may not need to input the whole table for decoding at all steps. As a preliminary study, we created a retrieval augmentation method that only includes the oracle objects (that are mentioned in the next step) as the input, but we see little improvement. We think more physical rules and math computation with the object features will help us gain more improvement. 
\end{itemize}


\bibliography{aaai23}

\begin{thebibliography}{31}
\providecommand{\natexlab}[1]{#1}

\bibitem[{Ahn et~al.(2022)Ahn, Brohan, Brown, Chebotar, Cortes, David, Finn,
  Gopalakrishnan, Hausman, Herzog, Ho, Hsu, Ibarz, Ichter, Irpan, Jang, Ruano,
  Jeffrey, Jesmonth, Joshi, Julian, Kalashnikov, Kuang, Lee, Levine, Lu, Luu,
  Parada, Pastor, Quiambao, Rao, Rettinghouse, Reyes, Sermanet, Sievers, Tan,
  Toshev, Vanhoucke, Xia, Xiao, Xu, Xu, and Yan}]{saycan2022arxiv}
Ahn, M.; Brohan, A.; Brown, N.; Chebotar, Y.; Cortes, O.; David, B.; Finn, C.;
  Gopalakrishnan, K.; Hausman, K.; Herzog, A.; Ho, D.; Hsu, J.; Ibarz, J.;
  Ichter, B.; Irpan, A.; Jang, E.; Ruano, R.~J.; Jeffrey, K.; Jesmonth, S.;
  Joshi, N.~J.; Julian, R.~C.; Kalashnikov, D.; Kuang, Y.; Lee, K.-H.; Levine,
  S.; Lu, Y.; Luu, L.; Parada, C.; Pastor, P.; Quiambao, J.; Rao, K.;
  Rettinghouse, J.; Reyes, D.~M.; Sermanet, P.; Sievers, N.; Tan, C.; Toshev,
  A.; Vanhoucke, V.; Xia, F.; Xiao, T.; Xu, P.; Xu, S.; and Yan, M. 2022.
\newblock Do As I Can, Not As I Say: Grounding Language in Robotic Affordances.
\newblock In \emph{Conference on Robot Learning}.

\bibitem[{Anderson et~al.(2016)Anderson, Fernando, Johnson, and
  Gould}]{Anderson2016SPICESP}
Anderson, P.; Fernando, B.; Johnson, M.; and Gould, S. 2016.
\newblock SPICE: Semantic Propositional Image Caption Evaluation.
\newblock In \emph{Proc. of ECCV}.

\bibitem[{Arik and Pfister(2021)}]{Arik2021TabNetAI}
Arik, S.~{\"{O}}.; and Pfister, T. 2021.
\newblock TabNet: Attentive Interpretable Tabular Learning.
\newblock In \emph{Thirty-Fifth {AAAI} Conference on Artificial Intelligence,
  {AAAI} 2021, Thirty-Third Conference on Innovative Applications of Artificial
  Intelligence, {IAAI} 2021, The Eleventh Symposium on Educational Advances in
  Artificial Intelligence, {EAAI} 2021, Virtual Event, February 2-9, 2021},
  6679--6687. {AAAI} Press.

\bibitem[{Brown et~al.(2020)Brown, Mann, Ryder, Subbiah, Kaplan, Dhariwal,
  Neelakantan, Shyam, Sastry, Askell, Agarwal, Herbert{-}Voss, Krueger,
  Henighan, Child, Ramesh, Ziegler, Wu, Winter, Hesse, Chen, Sigler, Litwin,
  Gray, Chess, Clark, Berner, McCandlish, Radford, Sutskever, and
  Amodei}]{gpt3}
Brown, T.~B.; Mann, B.; Ryder, N.; Subbiah, M.; Kaplan, J.; Dhariwal, P.;
  Neelakantan, A.; Shyam, P.; Sastry, G.; Askell, A.; Agarwal, S.;
  Herbert{-}Voss, A.; Krueger, G.; Henighan, T.; Child, R.; Ramesh, A.;
  Ziegler, D.~M.; Wu, J.; Winter, C.; Hesse, C.; Chen, M.; Sigler, E.; Litwin,
  M.; Gray, S.; Chess, B.; Clark, J.; Berner, C.; McCandlish, S.; Radford, A.;
  Sutskever, I.; and Amodei, D. 2020.
\newblock Language Models are Few-Shot Learners.
\newblock In Larochelle, H.; Ranzato, M.; Hadsell, R.; Balcan, M.; and Lin, H.,
  eds., \emph{Advances in Neural Information Processing Systems 33: Annual
  Conference on Neural Information Processing Systems 2020, NeurIPS 2020,
  December 6-12, 2020, virtual}.

\bibitem[{Chen et~al.(2021)Chen, Lu, Rajeswaran, Lee, Grover, Laskin, Abbeel,
  Srinivas, and Mordatch}]{Chen2021DecisionTR}
Chen, L.; Lu, K.; Rajeswaran, A.; Lee, K.; Grover, A.; Laskin, M.; Abbeel, P.;
  Srinivas, A.; and Mordatch, I. 2021.
\newblock Decision Transformer: Reinforcement Learning via Sequence Modeling.
\newblock In Ranzato, M.; Beygelzimer, A.; Dauphin, Y.~N.; Liang, P.; and
  Vaughan, J.~W., eds., \emph{Advances in Neural Information Processing Systems
  34: Annual Conference on Neural Information Processing Systems 2021, NeurIPS
  2021, December 6-14, 2021, virtual}, 15084--15097.

\bibitem[{Chen et~al.(2020)Chen, Wang, Chen, Zhang, Wang, Li, Zhou, and
  Wang}]{Chen2020TabFact}
Chen, W.; Wang, H.; Chen, J.; Zhang, Y.; Wang, H.; Li, S.; Zhou, X.; and Wang,
  W.~Y. 2020.
\newblock TabFact: {A} Large-scale Dataset for Table-based Fact Verification.
\newblock In \emph{Proc. of ICLR}. OpenReview.net.

\bibitem[{Farag and Saleh(2018)}]{Farag2018BehaviorCF}
Farag, W.~A.; and Saleh, Z. 2018.
\newblock Behavior Cloning for Autonomous Driving using Convolutional Neural
  Networks.
\newblock \emph{2018 International Conference on Innovation and Intelligence
  for Informatics, Computing, and Technologies (3ICT)}.

\bibitem[{Herzig et~al.(2020)Herzig, Nowak, M{\"u}ller, Piccinno, and
  Eisenschlos}]{herzig-etal-2020-tapas}
Herzig, J.; Nowak, P.~K.; M{\"u}ller, T.; Piccinno, F.; and Eisenschlos, J.
  2020.
\newblock {T}a{P}as: Weakly Supervised Table Parsing via Pre-training.
\newblock In \emph{Proc. of ACL}, 4320--4333. Online: Association for
  Computational Linguistics.

\bibitem[{Huang et~al.(2022)Huang, Abbeel, Pathak, and
  Mordatch}]{Huang2022LanguageMA}
Huang, W.; Abbeel, P.; Pathak, D.; and Mordatch, I. 2022.
\newblock Language Models as Zero-Shot Planners: Extracting Actionable
  Knowledge for Embodied Agents.
\newblock In Chaudhuri, K.; Jegelka, S.; Song, L.; Szepesv{\'{a}}ri, C.; Niu,
  G.; and Sabato, S., eds., \emph{International Conference on Machine Learning,
  {ICML} 2022, 17-23 July 2022, Baltimore, Maryland, {USA}}, volume 162 of
  \emph{Proceedings of Machine Learning Research}, 9118--9147. {PMLR}.

\bibitem[{Kolve et~al.(2017)Kolve, Mottaghi, Han, VanderBilt, Weihs, Herrasti,
  Deitke, Ehsani, Gordon, Zhu, Kembhavi, Gupta, and Farhadi}]{ai2thor}
Kolve, E.; Mottaghi, R.; Han, W.; VanderBilt, E.; Weihs, L.; Herrasti, A.;
  Deitke, M.; Ehsani, K.; Gordon, D.; Zhu, Y.; Kembhavi, A.; Gupta, A.~K.; and
  Farhadi, A. 2017.
\newblock AI2-THOR: An Interactive 3D Environment for Visual AI.
\newblock \emph{ArXiv preprint}, abs/1712.05474.

\bibitem[{Lewis et~al.(2020)Lewis, Liu, Goyal, Ghazvininejad, Mohamed, Levy,
  Stoyanov, and Zettlemoyer}]{Lewis2020BARTDS}
Lewis, M.; Liu, Y.; Goyal, N.; Ghazvininejad, M.; Mohamed, A.; Levy, O.;
  Stoyanov, V.; and Zettlemoyer, L. 2020.
\newblock {BART}: Denoising Sequence-to-Sequence Pre-training for Natural
  Language Generation, Translation, and Comprehension.
\newblock In \emph{Proc. of ACL}, 7871--7880. Online: Association for
  Computational Linguistics.

\bibitem[{Lin et~al.(2020)Lin, Zhou, Shen, Zhou, Bhagavatula, Choi, and
  Ren}]{Lin2020CommonGenAC}
Lin, B.~Y.; Zhou, W.; Shen, M.; Zhou, P.; Bhagavatula, C.; Choi, Y.; and Ren,
  X. 2020.
\newblock {C}ommon{G}en: A Constrained Text Generation Challenge for Generative
  Commonsense Reasoning.
\newblock In \emph{Findings of the Association for Computational Linguistics:
  EMNLP 2020}, 1823--1840. Online: Association for Computational Linguistics.

\bibitem[{Lin(2004)}]{Lin2004ROUGEAP}
Lin, C.-Y. 2004.
\newblock {ROUGE}: A Package for Automatic Evaluation of Summaries.
\newblock In \emph{Text Summarization Branches Out}, 74--81. Barcelona, Spain:
  Association for Computational Linguistics.

\bibitem[{Liu et~al.(2022{\natexlab{a}})Liu, Liu, He, and Yang}]{Liu2022LEBPL}
Liu, H.; Liu, Y.; He, H.; and Yang, H. 2022{\natexlab{a}}.
\newblock LEBP - Language Expectation \& Binding Policy: A Two-Stream Framework
  for Embodied Vision-and-Language Interaction Task Learning Agents.
\newblock \emph{ArXiv preprint}, abs/2203.04637.

\bibitem[{Liu et~al.(2022{\natexlab{b}})Liu, Chen, Guo, Ziyadi, Lin, Chen, and
  Lou}]{Liu2021TAPEXTP}
Liu, Q.; Chen, B.; Guo, J.; Ziyadi, M.; Lin, Z.; Chen, W.; and Lou, J.
  2022{\natexlab{b}}.
\newblock {TAPEX:} Table Pre-training via Learning a Neural {SQL} Executor.
\newblock In \emph{Proc. of ICLR}. OpenReview.net.

\bibitem[{Min et~al.(2022)Min, Chaplot, Ravikumar, Bisk, and
  Salakhutdinov}]{Min2021FILMFI}
Min, S.~Y.; Chaplot, D.~S.; Ravikumar, P.~K.; Bisk, Y.; and Salakhutdinov, R.
  2022.
\newblock {FILM:} Following Instructions in Language with Modular Methods.
\newblock In \emph{Proc. of ICLR}. OpenReview.net.

\bibitem[{Papineni et~al.(2002)Papineni, Roukos, Ward, and
  Zhu}]{Papineni2002BleuAM}
Papineni, K.; Roukos, S.; Ward, T.; and Zhu, W.-J. 2002.
\newblock {B}leu: a Method for Automatic Evaluation of Machine Translation.
\newblock In \emph{Proc. of ACL}, 311--318. Philadelphia, Pennsylvania, USA:
  Association for Computational Linguistics.

\bibitem[{Pashakhanloo et~al.(2022)Pashakhanloo, Naik, Wang, Dai, Maniatis, and
  Naik}]{pashakhanloo2022codetrek}
Pashakhanloo, P.; Naik, A.; Wang, Y.; Dai, H.; Maniatis, P.; and Naik, M. 2022.
\newblock CodeTrek: Flexible Modeling of Code using an Extensible Relational
  Representation.
\newblock In \emph{Proc. of ICLR}. OpenReview.net.

\bibitem[{Pashevich, Schmid, and Sun(2021)}]{Pashevich2021EpisodicTF}
Pashevich, A.; Schmid, C.; and Sun, C. 2021.
\newblock Episodic Transformer for Vision-and-Language Navigation.
\newblock In \emph{2021 {IEEE/CVF} International Conference on Computer Vision,
  {ICCV} 2021, Montreal, QC, Canada, October 10-17, 2021}, 15922--15932.
  {IEEE}.

\bibitem[{Pi et~al.(2022)Pi, Liu, Chen, Ziyadi, Lin, Fu, Gao, Lou, and
  Chen}]{Pi2022ReasoningLP}
Pi, X.; Liu, Q.; Chen, B.; Ziyadi, M.; Lin, Z.; Fu, Q.; Gao, Y.; Lou, J.-G.;
  and Chen, W. 2022.
\newblock Reasoning Like Program Executors.
\newblock In \emph{Proc. of EMNLP}, 761--779. Abu Dhabi, United Arab Emirates:
  Association for Computational Linguistics.

\bibitem[{Raffel et~al.(2020)Raffel, Shazeer, Roberts, Lee, Narang, Matena,
  Zhou, Li, and Liu}]{Raffel2020ExploringTL}
Raffel, C.; Shazeer, N.; Roberts, A.; Lee, K.; Narang, S.; Matena, M.; Zhou,
  Y.; Li, W.; and Liu, P.~J. 2020.
\newblock Exploring the Limits of Transfer Learning with a Unified Text-to-Text
  Transformer.
\newblock \emph{J. Mach. Learn. Res.}, 21: 140:1--140:67.

\bibitem[{Shridhar et~al.(2020)Shridhar, Thomason, Gordon, Bisk, Han, Mottaghi,
  Zettlemoyer, and Fox}]{ALFRED20}
Shridhar, M.; Thomason, J.; Gordon, D.; Bisk, Y.; Han, W.; Mottaghi, R.;
  Zettlemoyer, L.; and Fox, D. 2020.
\newblock {ALFRED:} {A} Benchmark for Interpreting Grounded Instructions for
  Everyday Tasks.
\newblock In \emph{2020 {IEEE/CVF} Conference on Computer Vision and Pattern
  Recognition, {CVPR} 2020, Seattle, WA, USA, June 13-19, 2020}, 10737--10746.
  {IEEE}.

\bibitem[{Shridhar et~al.(2021)Shridhar, Yuan, C{\^{o}}t{\'{e}}, Bisk,
  Trischler, and Hausknecht}]{Shridhar2021ALFWorldAT}
Shridhar, M.; Yuan, X.; C{\^{o}}t{\'{e}}, M.; Bisk, Y.; Trischler, A.; and
  Hausknecht, M.~J. 2021.
\newblock ALFWorld: Aligning Text and Embodied Environments for Interactive
  Learning.
\newblock In \emph{Proc. of ICLR}. OpenReview.net.

\bibitem[{Talmor et~al.(2019)Talmor, Herzig, Lourie, and
  Berant}]{Talmor2019CommonsenseQAAQ}
Talmor, A.; Herzig, J.; Lourie, N.; and Berant, J. 2019.
\newblock {C}ommonsense{QA}: A Question Answering Challenge Targeting
  Commonsense Knowledge.
\newblock In \emph{Proc. of NAACL-HLT}, 4149--4158. Minneapolis, Minnesota:
  Association for Computational Linguistics.

\bibitem[{Vedantam, Zitnick, and Parikh(2015)}]{Vedantam2015CIDErCI}
Vedantam, R.; Zitnick, C.~L.; and Parikh, D. 2015.
\newblock CIDEr: Consensus-based image description evaluation.
\newblock In \emph{{IEEE} Conference on Computer Vision and Pattern
  Recognition, {CVPR} 2015, Boston, MA, USA, June 7-12, 2015}, 4566--4575.
  {IEEE} Computer Society.

\bibitem[{Wang et~al.(2022)Wang, Jansen, C{\^o}t{\'e}, and
  Ammanabrolu}]{scienceworld2022}
Wang, R.; Jansen, P.; C{\^o}t{\'e}, M.-A.; and Ammanabrolu, P. 2022.
\newblock {S}cience{W}orld: Is your Agent Smarter than a 5th Grader?
\newblock In \emph{Proc. of EMNLP}, 11279--11298. Abu Dhabi, United Arab
  Emirates: Association for Computational Linguistics.

\bibitem[{Wang et~al.(2019)Wang, Wu, Chen, Li, Wang, and
  Wang}]{Wang2019VaTeXAL}
Wang, X.; Wu, J.; Chen, J.; Li, L.; Wang, Y.; and Wang, W.~Y. 2019.
\newblock VaTeX: {A} Large-Scale, High-Quality Multilingual Dataset for
  Video-and-Language Research.
\newblock In \emph{2019 {IEEE/CVF} International Conference on Computer Vision,
  {ICCV} 2019, Seoul, Korea (South), October 27 - November 2, 2019},
  4580--4590. {IEEE}.

\bibitem[{Yi et~al.(2018)Yi, Wu, Gan, Torralba, Kohli, and
  Tenenbaum}]{vqa2018nips}
Yi, K.; Wu, J.; Gan, C.; Torralba, A.; Kohli, P.; and Tenenbaum, J. 2018.
\newblock Neural-Symbolic {VQA:} Disentangling Reasoning from Vision and
  Language Understanding.
\newblock In Bengio, S.; Wallach, H.~M.; Larochelle, H.; Grauman, K.;
  Cesa{-}Bianchi, N.; and Garnett, R., eds., \emph{Advances in Neural
  Information Processing Systems 31: Annual Conference on Neural Information
  Processing Systems 2018, NeurIPS 2018, December 3-8, 2018, Montr{\'{e}}al,
  Canada}, 1039--1050.

\bibitem[{Yin et~al.(2020)Yin, Neubig, Yih, and Riedel}]{yin-etal-2020-tabert}
Yin, P.; Neubig, G.; Yih, W.-t.; and Riedel, S. 2020.
\newblock {T}a{BERT}: Pretraining for Joint Understanding of Textual and
  Tabular Data.
\newblock In \emph{Proc. of ACL}, 8413--8426. Online: Association for
  Computational Linguistics.

\bibitem[{Yoran, Talmor, and Berant(2022)}]{yoran-etal-2022-turning}
Yoran, O.; Talmor, A.; and Berant, J. 2022.
\newblock Turning Tables: Generating Examples from Semi-structured Tables for
  Endowing Language Models with Reasoning Skills.
\newblock In \emph{Proc. of ACL}, 6016--6031. Dublin, Ireland: Association for
  Computational Linguistics.

\bibitem[{Zhang et~al.(2020)Zhang, Kishore, Wu, Weinberger, and
  Artzi}]{Zhang2020BERTScoreET}
Zhang, T.; Kishore, V.; Wu, F.; Weinberger, K.~Q.; and Artzi, Y. 2020.
\newblock BERTScore: Evaluating Text Generation with {BERT}.
\newblock In \emph{Proc. of ICLR}. OpenReview.net.

\end{thebibliography}

\end{document}